\documentclass{article}

\usepackage{microtype}
\usepackage{times}
\usepackage{graphicx}
\usepackage{amsmath}
\usepackage{amssymb}
\usepackage{bm}
\usepackage{subfigure}
\usepackage{color}
\usepackage{url}
\usepackage{pdfsync}
\usepackage{booktabs} 

\usepackage{hyperref}

\def\etal{\emph{et al.}}
\def\ie{\emph{i.e.}}

\graphicspath{{Figs/}}

\usepackage[accepted]{icml2018}

\begin{document}

\twocolumn[
\icmltitle{Latent Dirichlet Allocation in Generative Adversarial Networks}



\icmlsetsymbol{equal}{*}

\begin{icmlauthorlist}
\icmlauthor{Lili Pan}{to}
\icmlauthor{Shen Cheng}{to}
\icmlauthor{Jian Liu}{to}
\icmlauthor{Yazhou Ren}{to}
\icmlauthor{Zenglin Xu}{to}
\end{icmlauthorlist}

\icmlaffiliation{to}{Statistical Machine Intelligence and Learning Lab, University of Electronic Science and Technology of China, Chengdu, China}

\icmlcorrespondingauthor{Lili Pan}{panlili8255@gmail.com}
\icmlcorrespondingauthor{Zenglin Xu}{zenglin@gmail.com}


\vskip 0.3in
]



\printAffiliationsAndNotice{}  

\begin{abstract}
We study the problem of multimodal generative modelling of images based on generative adversarial networks (GANs).
Despite the success of existing methods, they often ignore the underlying structure of vision data or its multimodal generation characteristics.
To address this problem, we introduce the Dirichlet prior for multimodal image generation, which leads to a new Latent Dirichlet Allocation based GAN (LDAGAN).
In detail, for the generative process modelling, LDAGAN defines a generative mode for each sample, determining which generative sub-process it belongs to.
For the adversarial training, LDAGAN derives a variational  expectation-maximization (VEM) algorithm to estimate model parameters.
Experimental results on real-world datasets have demonstrated the outstanding performance of LDAGAN over other existing GANs.
\end{abstract}

\section{Introduction}
Generating realistic images has been actively pursued in the machine learning community in recent years. Achieving this goal requires true understanding of images, including the structure, semantics and so on.
Deep generative models (DGMs)~\cite{goodfellow2014generative,diederik2014vae} have attracted considerable attention recently because they provide us a new perspective to deeply understand vision data.
Among various DGMs, generative adversarial networks (GANs)~\cite{goodfellow2014generative} have gained the most interest as they learn a deep generative model for which no explicit likelihood function is required, but only a generative process~\cite{uehara2016generative,mohamed2017learning,nowozin2016f,tran2017deep}.

Towards the goal of generating realistic images, various GANs have been proposed with varying degrees of success~\cite{salimans2016improved,nguyen2017dual,miyato2018spectral,tran2018dist,tolstikhin2017adagan}.
Most of them are non-structured and can be roughly categorized into two categories: single-generator based and multi-generator based, depending on the number of generators employed.
Single-generator based approaches, for example~\cite{nguyen2017dual,miyato2018spectral,salimans2016improved,tran2018dist}, try to modify the objective of GANs, or the optimization strategies to guide the training process.
Multiple-generator based approaches, for example, the work proposed by~\cite{hoang2018mgan,tolstikhin2017adagan,arora2017generalization}, employ multiple generators to generate more diverse images.
However, both categories of approaches ignore that realistic generation essentially depends on truly understanding data, especially the structure.

A meaningful step forward in the regard was made by the Graphical GAN (GMGAN)~\cite{NIPS2018_7846}, which employs Bayesian networks to model the structured generative process of images. However, GMGAN only defines a single generative process (\ie~generator) transforming from mixture of Gaussian noise to images.
In fact, real-world images, such as images in the CIFAR-10 and ImageNet datasets, are highly complex and usually have multi-modality.
For such complex data, a single generative process is almost impossible to fit for all images, resulting in problems like mode collapse and dropping~\cite{hoang2018mgan,tolstikhin2017adagan,arora2017generalization}.

\begin{figure}
 \centering
    \includegraphics[width=0.22\textwidth]{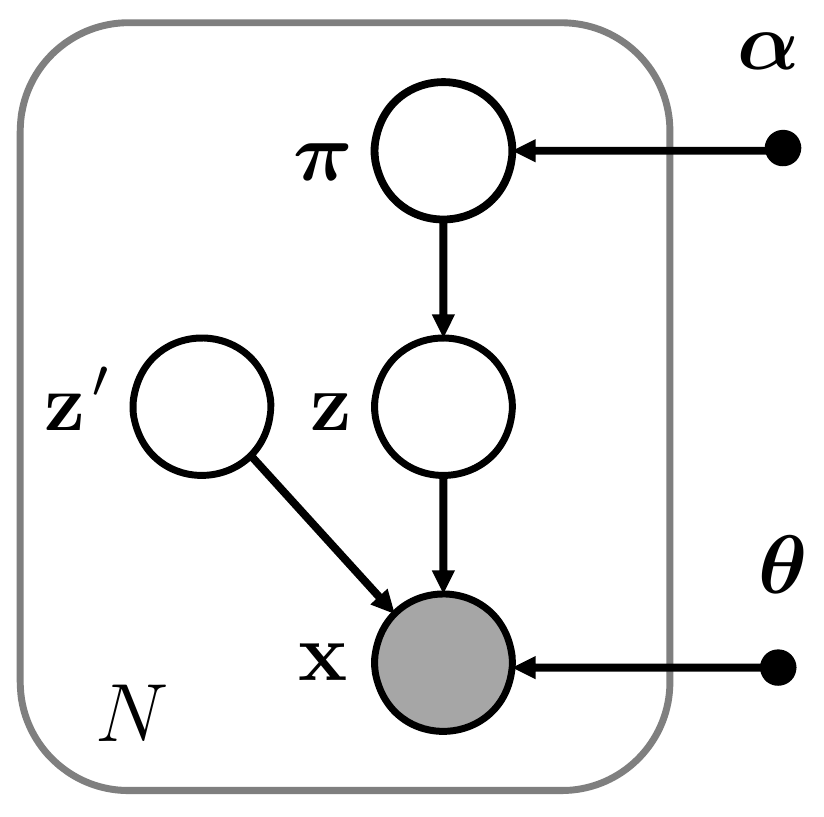}
     \caption{Graphical model for LDAGAN. It is a three-level hierarchical deep Bayesian model. Latent variables $\bm{\pi}$ describe mode distribution which have a Dirichlet prior $\text{Dir}(\bm{\alpha})$. $\mathbf{z}'$ and $\mathbf{z}$ represent the input noise variables and latent mode of each sample respectively. The parameters $\bm{\theta}=\left\{\bm{\theta}_1,...,\bm{\theta}_K\right\}$ are associated with $K$ generators $G=\left\{G_1,...,G_K\right\}$. }
     \label{fig:LDA-GAN}
\end{figure}

To address these issues, we propose a multi-modal generative process for images based on GANs and use a probabilistic graphical model to represent the generation process, as illustrated in
Fig.~\ref{fig:LDA-GAN}. 
Our key idea lies in introducing an underlying generative mode $\mathbf{z}$ for each sample, denoting which generative sub-process (\ie~generator $G_k$) it belongs to.
To achieve more precise representation capability, we suppose the mode distribution $\bm{\pi}$ could be distinct for each sample, while sampled from the same Dirichlet prior distribution~\cite{blei2003latent}.
This line of thinking leads to a new structured and implicit generative model: latent Dirichlet allocation based GANs (LDAGAN), which not only the natural multi-modality structure of image, but also can be more interpretable as a topic model~\cite{blei2003latent}.

Given the strong representation power of LDAGAN, a natural question then arises: \emph{how to make inference and estimate model parameters in such a three-level hierarchical deep Bayesian networks without explicit likelihood?}
To this end, we take an important step by presenting a variational inference and expectation-maximization (EM) algorithm in an adversarial process.
Specifically, we utilize the discriminator in GANs to formulate the likelihood function for model parameters.
In adversarial training, we maximize the above likelihood with respect to model parameters by virtue of EM algorithm.
We make stochastic variation inference to ensure the training of LDAGAN is not time consuming.

The main contributions of this paper are summarized as follows:
(\romannumeral1) We build a structured GANs exploring multimodal generative process of images.
(\romannumeral2) We present a variational EM algorithm for Bayesian network parameter estimation in adversarial training.
(\romannumeral3) We achieve state-of-the-art performance on CIFAR-10~\cite{krizhevsky2009learning}, CIFAR-100~\cite{krizhevsky2009learning} and ImageNet~\cite{russakovsky2015imagenet} datasets. For example, our method has achieved a value of $28.9$ for Fr\'{e}chet Inception Distance on the ImageNet dataset, which is currently the best reported result with standard CNN architecture in literature to our knowledge.

\section{Previous Work}
\label{sec:Previous}

This section reviews previous work on variants of generative adversarial networks (GANs).

\subsection{No Structured GANs}
\label{sec:GANs}

\textbf{Single-Generator based GANs:} Generative Adversarial Networks learn a generator that transforms input noise variables to target distribution.
Conventionally, the generator is a nonlinear function, specified by a deep network.
One main difficulty in training GANs is how to avoid mode collapse and dropping whilst affording an efficient evaluation.
To account for this, many important variants of GANs have been proposed.
Salimans \etal~\cite{salimans2016improved} introduced several techniques into GANs training, including feature matching, minibatch discrimination, historical averaging and virtual batch normalization, to avoid mode collapse.
WGAN leverages a smooth metric (\ie~Wasserstein distance) to measure the distance between two probability distributions to improve the stablity of GANs training~\cite{arjovsky2015wasserstein}.
WGAN-GP replaces the weight clipping in WGAN with penalizing the gradient norm of the interpolated samples to achieve more stable performance~\cite{gulrajani2017improved}.
WGAN-GP+TURR uses a two time-scale update rule for training GANs to guarantee it converges to a stationary local Nash equilibrium~\cite{heusel2017gans}.
Recently, Miyato \etal~\cite{miyato2018spectral} have applied spectral normalization to stabilize training the discriminator of SNGANs, rendering generated samples more diverse.

Although these improvements in GANs are effective somewhat, their performance tends to be unsatisfactory when real data are highly complex.
In fact, a single generator is hard to properly capture complex image generation process, especially when the images are obviously multimodal.
The low model capacity may be one of the primary reasons to render mode collapse/dropping.

\textbf{Multi-Generator based GANs:} To account for the drawbacks of single generator based GANs, in~\cite{hoang2018mgan,tolstikhin2017adagan,arora2017generalization}, multi-generator based GANs train multiple generators to capture more modes of data.
For example, AdaGAN and boosting-inspired GANs learn to generate samples of some modes with one generator and then remove samples of the same modes in training set to train a next generator~\cite{tolstikhin2017adagan}.
To simplify this procedure, Mix-GAN and MGAN suppose all generators together induce a mixture of sub-modal distributions~\cite{hoang2018mgan,arora2017generalization}, leading to a more complex and flexible model distribution.
This permits model distribution to become more close to complex real distribution.

Although multi-generator based GANs seem to be capable of generate more diverse samples, they exhibit two drawbacks.
Firstly, the underlying structure of data are not explored and represented in GANs, for example in MGAN and MixGAN~\cite{hoang2018mgan,arora2017generalization}.
A simple mixing scheme appears to provide no guarantee that the model distribution is able to cover all modes of images.
Secondly, without no structure information, some multi-generator based GANs~\cite{hoang2018mgan} encourage mode diversity of generated samples, resulting in intra-class mode dropping.

\subsection{Structured GANs}
\label{sec:GraGAN}

To overcome the drawbacks of no structured GANs, in~\cite{NIPS2018_7846}, Graphical-GAN uses Bayesian networks to represent the structure of vision data and conjoins GANs to generate images.
This hints us probabilistic graphical model could be used in GANs to model the generative process.
Moreover, some approximate inference and learning algorithms have been proposed for them~\cite{karaletsos2016advermp}

Although structured GANs take into consideration the underlying structure of data in generation, they appear to have not exactly model the multimodal generation process of images yet, since they have relatively low inception score~\cite{NIPS2018_7846}.

\section{Latent Dirichlet Allocation in GANs}
\label{sec:LDAGAN}

To clearly describe the multimodal generation process of images, we define the following image generation process:
\begin{enumerate}
\item Choose mode distribution $\bm{\pi}\sim\text{Dir}(\bm{\alpha})$.
\item Choose a mode $\mathbf{z}\sim\text{Mult}(\bm{\pi})$.
\item Generate a sample $\mathbf{x}$ conditioned on noise $\mathbf{z}'$ and mode $\mathbf{z}$, that is, $\mathbf{x}=G_{k}\left(\mathbf{z}'; \bm{\theta}_k\right)$ if $z_k=1$.
\end{enumerate}
The graphical model representing this process is illustrated in Fig.~\ref{fig:LDA-GAN}.
Here, $z_k=1$ indicates which mode was chosen for generating sample $\mathbf{x}$, where $z_k\in\left\{0,1\right\}$ and $\sum_k z_k=1$.
That is, for each input noise $\mathbf{z}'$, we choose a mode specific generator $G_k$ parameterized by $\bm{\theta}_k$ for generation.

The objective for learning $\bm{\theta}$ and $\bm{\alpha}$ is not explicit as there is no likelihood function being specified, only a generating process.
However, adversarial learning opens a door to solve this problem.
$\bm{\theta}$ and $\bm{\alpha}$ can be optimized through minimizing the divergence between model distribution $p_g\left(\mathbf{x}\right)$ and real distributions $p_{data}\left(\mathbf{x}\right)$.
In the work~\cite{mohamed2017learning}, Mohamed \etal~revealed training a discriminator in GANs is equivalent to training a good estimator to measure the distance between the two distributions.

Following this, we learn a discriminator $D\left(\mathbf{x}; \bm{\phi}\right)$, a function bounded in $\left[0,1\right]$ with parameters $\bm{\phi}$, to output the probability of $\mathbf{x}$ belonging to real data, denoted by $p\left(y=1|\mathbf{x},\bm{\phi}\right)$.
Here, the binary variable $y$ indicates whether $\mathbf{x}$ is real or fake.
We hope $D$ to maximize the probability of assigning the correct label to both the real samples and samples generated from $\mathbf{z}'$.
Meanwhile, we hope $\bm{\alpha}$ and $\bm{\theta}$ to minimize $\log \left( 1-p\left( y=1 | \mathbf{z}',\bm{\theta},\bm{\alpha},\bm{\phi} \right)\right)$.
Then, the objective function of LDAGAN takes the form:
\begin{multline}
\min_{\bm{\theta}, \bm{\alpha}}
\max_{\bm{\phi}}
\mathbb{E}_{\mathbf{x} \sim p_{data}\left(\mathbf{x}\right)}
\left[ \log p\left( y=1 | \mathbf{x}, \bm{\phi} \right) \right] \\
+\mathbb{E}_{\mathbf{z}' \sim p\left(\mathbf{z}'\right)}
\left[ \log \left( 1-p\left( y=1 | \mathbf{z}',\bm{\theta},\bm{\alpha},\bm{\phi} \right)\right) \right].
\label{eq:LDAGAN3}
\end{multline}
\noindent Compared with traditional GANs, LDAGAN formulates its objective function in a probabilistic form.

\section{Learning}
\label{sec:Learning}

This section describes the learning of the discriminator, generators and Dirichlet parameters $\bm{\alpha}$ in LDAGAN.
\subsection{Discriminators}
\label{sec:Discriminator}

We extract the terms only containing $\bm{\phi}$ from Eq.~(\ref{eq:LDAGAN3}) to construct the discriminative loss:
\begin{multline}
\label{eq:Learning-D1}
\max_{\bm{\phi}}
\mathbb{E}_{\mathbf{x} \sim p_{data}\left(\mathbf{x}\right)}
\mathbb{E}\left[ \log p\left( y=1 | \mathbf{x},\bm{\phi} \right) \right]  \\
+\mathbb{E}_{\mathbf{z}' \sim p\left(\mathbf{z}'\right)}
\mathbb{E}\left[ \log\left( 1-p\left( y=1 | \mathbf{z}',\bm{\theta},\bm{\alpha},\bm{\phi} \right) \right) \right],
\end{multline}
where $p\left( y=1 | \mathbf{z}',\bm{\theta},\bm{\alpha},\bm{\phi} \right)$ is a marginal probability, representing the probability of the sample generated from $\mathbf{z}'$ being real.
It is obtained by integrating joint distribution over $\bm{\pi}$ and summing over $\mathbf{z}$:
\begin{equation}
\int p\left(\bm{\pi} | \bm{\alpha}\right)
\left(\sum_{\mathbf{z}}
p\left(\mathbf{z}|\bm{\pi}\right)
p\left(y=1|\mathbf{z}, \mathbf{z}', \bm{\theta}, \bm{\phi}\right)
\right)d\bm{\pi}.
\label{eq:Learning-D2}
\end{equation}
Obviously, the joint distribution is given as a product of conditionals in the form $p\left(\bm{\pi} | \bm{\alpha}\right)p\left(\mathbf{z}|\bm{\pi}\right)p\left(y=1|\mathbf{z}, \mathbf{z}', \bm{\theta}, \bm{\phi}\right)$.
$p\left(y=1|\mathbf{z}, \mathbf{z}', \bm{\theta}, \bm{\phi}\right)$ here denotes, given the underlying mode (\ie~$z_k=1$), the probability of the generated sample being real.
We utilize $ D\left(G_k\left(\mathbf{z}'\right)\right)$ to score this probability.
Substituting Eq.~(\ref{eq:Learning-D2}) back into the discriminative loss, the parameters $\bm{\phi}$ can be optimized after sampling since the integration over $\bm{\pi}$ is analytically intractable.
One note that despite we employs $K$ generators in our LDAGAN, we use only one discriminator.
This is because one discriminator helps to keep the balance of generators in training.

\subsection{Generators and Dirichlet parameters}
In this section, we turn to the optimization of generators and the Dirichlet parameters $\bm{\alpha}$.

\subsubsection{Generative Loss}
\label{sec:GeLoss}

Considering the terms related to generation in Eq.~(\ref{eq:LDAGAN3}), minimizing $\log\left(1-p\left(y=1|\mathbf{z}',\bm{\theta},\bm{\alpha}, \bm{\phi}\right)\right)$ with respect to $\bm{\theta}$ and $\bm{\alpha}$ is equivalent to maximizing $\log p\left(y=1|\mathbf{z}',\bm{\theta},\bm{\alpha}, \bm{\phi}\right)$ with respect to them.
Then, we rewrite the generative loss in the form:
\begin{equation}
\max_{\bm{\theta},\bm{\alpha}}
\mathbb{E}_{\mathbf{z}' \sim p \left( \mathbf{z}' \right)}
\left[ \log p\left( y=1 | \mathbf{z}',\bm{\theta},\bm{\alpha}, \bm{\phi} \right) \right],
\label{eq:Learning-G1}
\end{equation}
which means maximizing the probability of the samples generated from $\mathbf{z}'$ being real.
One note this probability is a likelihood function for model parameters in LDAGAN.
The maximum can be achieved if, and only if, our generative model finds the underlying structure of real data correctly and each generator models data of each mode appropriately.
In fact, maximizing the above likelihood is nontrivial for two reason:
(\romannumeral1) it is a deep likelihood, and
(\romannumeral2) it includes discrete and continuous latent variables $\mathbf{z}$ and $\bm{\pi}$.
In such a case, we theoretically propose to \emph{use the so called variational EM algorithm to maximize it}.

\subsubsection{Variational EM Algorithm}
\label{sec:EM}

The EM algorithm provides an useful way to find maximum likelihood solutions for probabilistic models having latent variables.
In general, it consists of three items, namely: variational distribution, E-step optimization and and M-step optimization.

\noindent\textbf{Variational Distribution:}
On the basis of the mean-field approximation, we define a variational distribution $q\left(\bm{\pi},\mathbf{z}|\bm{\gamma},\bm{\omega}\right)$ over the latent variables $\bm{\pi}$ and $\bm{z}$:
\begin{equation}
\label{eq:Learning-G2}
q\left(\bm{\pi}, \mathbf{z}|\bm{\gamma},\bm{\omega}\right)
= q\left(\bm{\pi}|\bm{\gamma}\right)
q\left(\mathbf{z}|\bm{\omega}\right),
\end{equation}
\noindent where $\bm{\gamma}$ is the Dirichlet parameters and $\bm{\omega}$ is the multinomial parameters.
Furthermore, we decompose the log likelihood function in Eq.~(\ref{eq:Learning-G1}) into the sum of lower bound function and KL divergence:
\begin{multline}
\log p\left(y=1|\mathbf{z}',\bm{\theta},\bm{\alpha}, \bm{\phi}\right)
=
L\left( \bm{\gamma},\bm{\omega};\bm{\alpha},\bm{\theta} \right)   \\
+ \text{KL}\left( q\left( \bm{\pi},\mathbf{z} | \bm{\gamma},\bm{\omega} \right)
|| p\left( \bm{\pi}, \mathbf{z}|y=1, \mathbf{z}',\bm{\theta},\bm{\alpha},\bm{\phi} \right) \right).
\label{eq:Learning-G3}
\end{multline}
Here $L\left( \bm{\gamma},\bm{\omega};\bm{\theta},\bm{\alpha} \right)$ is the lower bound function on $\log p\left(y=1|\mathbf{z}',\bm{\theta},\bm{\alpha}, \bm{\phi}\right)$, which is a function of variational parameters $\bm{\gamma}$ and $\bm{\omega}$, and also a function of the parameters $\bm{\theta}$ and $\bm{\alpha}$.
The detailed derivation of this decomposition can be found in Appendix A.
It is easily verified that the lower bound $L\left( \bm{\gamma},\bm{\omega};\bm{\theta},\bm{\alpha} \right)$ is maximized when the KL divergence vanishes.
From this perspective, according to Eq.~(\ref{eq:Learning-G3}), the variational distribution can be viewed as an approximation to the posterior distribution $p\left( \bm{\pi}, \mathbf{z}|y=1, \mathbf{z}',\bm{\theta},\bm{\alpha},\bm{\phi} \right)$.

\noindent\textbf{E-Step Optimization:}
The variational EM algorithm is a two-stage iterative optimization algorithm.
The E-step involves maximizing the lower bound with respect to the variational parameters $\bm{\gamma}$ and $\bm{\omega}$.
When model parameters $\bm{\theta}$ and $\bm{\alpha}$ are fixed, computing the derivatives of $L\left( \bm{\gamma},\bm{\omega};\bm{\theta},\bm{\alpha} \right)$ with respect to $\omega_k$ and $\gamma_k$, and setting them equal to zeros yields the following formulations for variational parameters updating (see Appendix A),
\begin{equation}
\label{eq:Learning-G4}
\omega_k\propto
D\left( G_k\left( \mathbf{z}'\right) \right)\exp\left( \Psi\left(\gamma_k\right) - \Psi \left( \sum\nolimits_{j=1}^K \gamma_j\right) \right),
\end{equation}
\begin{equation}
\label{eq:Learning-G5}
\gamma_k
=
\alpha_k
+
\omega_k,
\end{equation}
where $\Psi\left(\cdot\right)$ is known as the digamma function, and $\omega_k$ is the $k^{th}$ elements of $\bm{\omega}$ that should be normalized to make $\sum_{k=1}^K\omega_k=1$.
Here, $D\left(G_k\left(\mathbf{z}'\right)\right)$ denotes the likelihood term, which reflects the probability of the sample generated from $\mathbf{z}'$, given underlying mode $\mathbf{z}$ (\ie~$z_k=1$), being real.
$\exp\left( \Psi\left(\gamma_k\right) - \Psi \left( \sum\nolimits_{j=1}^K \gamma_j\right)\right)$ is related to the prior, where $\Psi\left(\gamma_k\right) - \Psi \left( \sum\nolimits_{j=1}^K \gamma_j \right) = \mathbb{E}_q\left[\log \pi_k \right]$.
As such, the multinomial update can be though of as a posterior multinomial according to Bayes' theorem.
Similarly, the Dirichlet update, shown in Eq.~(\ref{eq:Learning-G5}), can be viewed as a posterior Drichlet.
Then, updating $\omega_k$ and $\gamma_k$ alternatively until some convergence criterion is met,
results in optimal $\bm{\omega}$ and $\bm{\gamma}$ which maximize the above lower bound.
One note that the variational parameters vary as a function of $\mathbf{z}'$, and thus we rewrite them in the form $\bm{\gamma}\left(\mathbf{z}'\right)$ and $\bm{\omega}\left(\mathbf{z}'\right)$.

\noindent\textbf{M-Step Optimization:} In the subsequent M-step, with $\bm{\gamma}$ and $\bm{\omega}$ fixed, we maximize the expected lower bound $\mathcal{L}\left( \bm{\gamma},\bm{\omega};\bm{\theta},\bm{\alpha} \right)$ with respect to $\bm{\theta}$ and $\bm{\alpha}$.
It should be emphasized that the symbol $\mathcal{L}$ here denotes the expectation of the lower bound $L$, that is, $L$ is only associated with one sample $\mathbf{z}'$, while $\mathcal{L}$ is the expectation averaging over all samples.

Maximizing $\mathcal{L}\left( \bm{\gamma},\bm{\omega};\bm{\theta},\bm{\alpha} \right)$ with respect to $\bm{\theta}_k$ yields:
\begin{equation}
\max_{\bm{\theta}_k}
\mathbb{E}_{\mathbf{z}' \sim p_{\mathbf{z}'}}
\left[
\omega_k\left( \mathbf{z}' \right)
\log D\left( G_k\left( \mathbf{z}' \right) \right)
\right],
\label{eq:Learning-G6}
\end{equation}
where $\bm{\theta}_k$ is the parameters of generator $G_k$, and $\omega_k\left( \mathbf{z}' \right)$ is a posterior approximation.
$\omega_k\left( \mathbf{z}' \right)$ approximates the posterior probability of generated sample, under the assumption of being real, being generated by $G_k$.
Therefore, an appealing intuitive explanation for Eq.~(\ref{eq:Learning-G6}) is that each sample, when optimizing $G_k$, is weighted, guiding each generator to give more considerations to the 'good' samples.
Moreover, $G_k$ share their parameters with each other except the first layer, largely reducing the parameter number and improving training efficiency.

\begin{algorithm}[t]
   \caption{Minibatch stochastic optimization for LDAGAN.}
   \label{alg:Learning}
\begin{algorithmic}[1]

   \REQUIRE Initialize $\bm{\theta}$, $\bm{\alpha}$, and $\bm{\phi}$.
   \FOR{number in training iterations}
   \STATE Sample $M$ generated examples $\left\{\mathbf{x}'_m\right\}_{m=1}^M$.
   \STATE Sample $M$ real examples $\left\{\mathbf{x}_m\right\}_{m=1}^M$.
   \STATE Update $D$ by ascending gradient:
   \STATE {$\nabla_{\bm{\phi}}\frac{1}{M}\sum_{m=1}^M\left\{ \log D\left(\mathbf{x}_m\right) + \log \left(1-D\left(\mathbf{x}'_m\right)\right)\right\}$}.
   \STATE Sample $M$ noise samples $\left\{\mathbf{z}'_m\right\}_{m=1}^M$.
   \REPEAT
      \STATE Calculate $\bm{\omega}\left(\mathbf{z}'_m\right)$ in Eq.~(\ref{eq:Learning-G4}), $m=1,..,M$.
      \STATE Calculate $\bm{\gamma}\left(\mathbf{z}'_m\right)$ in Eq.~(\ref{eq:Learning-G5}), $m=1,..,M$.
   \UNTIL $\bm{\omega}\left(\mathbf{z}'_m\right)$ and $\bm{\gamma}\left(\mathbf{z}'_m\right)$ converge
   \STATE Update $\left\{G_k\right\}_{k=1}^K$ by ascending gradient:
   \STATE{$\nabla_{\bm{\theta}_k}\frac{1}{M}\sum_{m=1}^M\left\{\omega_k\left(\mathbf{z}'_m \right) \log D\left( G_k\left( \mathbf{z}'_m \right)\right)\right\}$}.
   \STATE Update $\bm{\alpha}$ by ascending gradient of Eq.~(\ref{eq:Learning-G7}).
   \ENDFOR
\end{algorithmic}
\end{algorithm}

The terms in $\mathcal{L}\left( \bm{\gamma},\bm{\omega};\bm{\theta},\bm{\alpha} \right)$ related to $\bm{\alpha}$ takes the form $\mathcal{L}_{\bm{\alpha}}$:
\begin{multline}
\log \Gamma \left( \sum\nolimits_{j=1}^K \alpha_j \right)
-\sum_{k=1}^K \log \Gamma \left( \alpha_k \right)
+
\mathbb{E}_{\mathbf{z}' \sim p\left(\mathbf{z}'\right)}\\
\left[
\sum_{k=1}^K
\left( \alpha_k-1 \right)
\left( \Psi\left(\gamma_k\left(\mathbf{z}'\right)\right)
-\Psi\left(\sum\nolimits_{j=1}^K\gamma_j\left(\mathbf{z}'\right)\right) \right)
\right].
\label{eq:Learning-G7}
\end{multline}
Taking the derivative with respect to $\bm{\alpha}$ and using gradient ascending, we will finally obtain the optimal solution for $\bm{\alpha}$.

\section{Stochastic Optimization}
\label{Stoch}

The learning of LDAGAN is a two step alternative optimization procedure: (\romannumeral1) update discriminator, and (\romannumeral2) use EM algorithm to maximize a deep likelihood with respect to $\bm{\theta}$ and $\bm{\alpha}$.
Such a procedure appears to be reasonable, but it is unclear how to incorporate the EM algorithm into this adversarial learning framework.
Inspired by stochastic variational inference~\cite{hoffman2013stochastic}, we propose to update $\bm{\theta}$ and $\bm{\alpha}$ as well as variational parameters based on only a subset of the training data, namely stochastic minibatch optimization.
At each iteration, we take only one step in the gradient direction.
The complete learning procedure is outlined in Algorithm~\ref{alg:Learning}.
This differs from standard GANs optimization only in that the variational parameters as well as Dirichlet parameters $\bm{\alpha}$ need to be updated at each iteration.
The related calculation is not time consuming, and hence the whole procedure should be fast.

\section{Experiments}
\label{sec:Experiments}
We carried out experiments on both synthetic data and real-world data.

\subsection{Evaluation on Synthetic Data}
\begin{figure}[ht]
  \centering
    \includegraphics[width=0.43\textwidth]{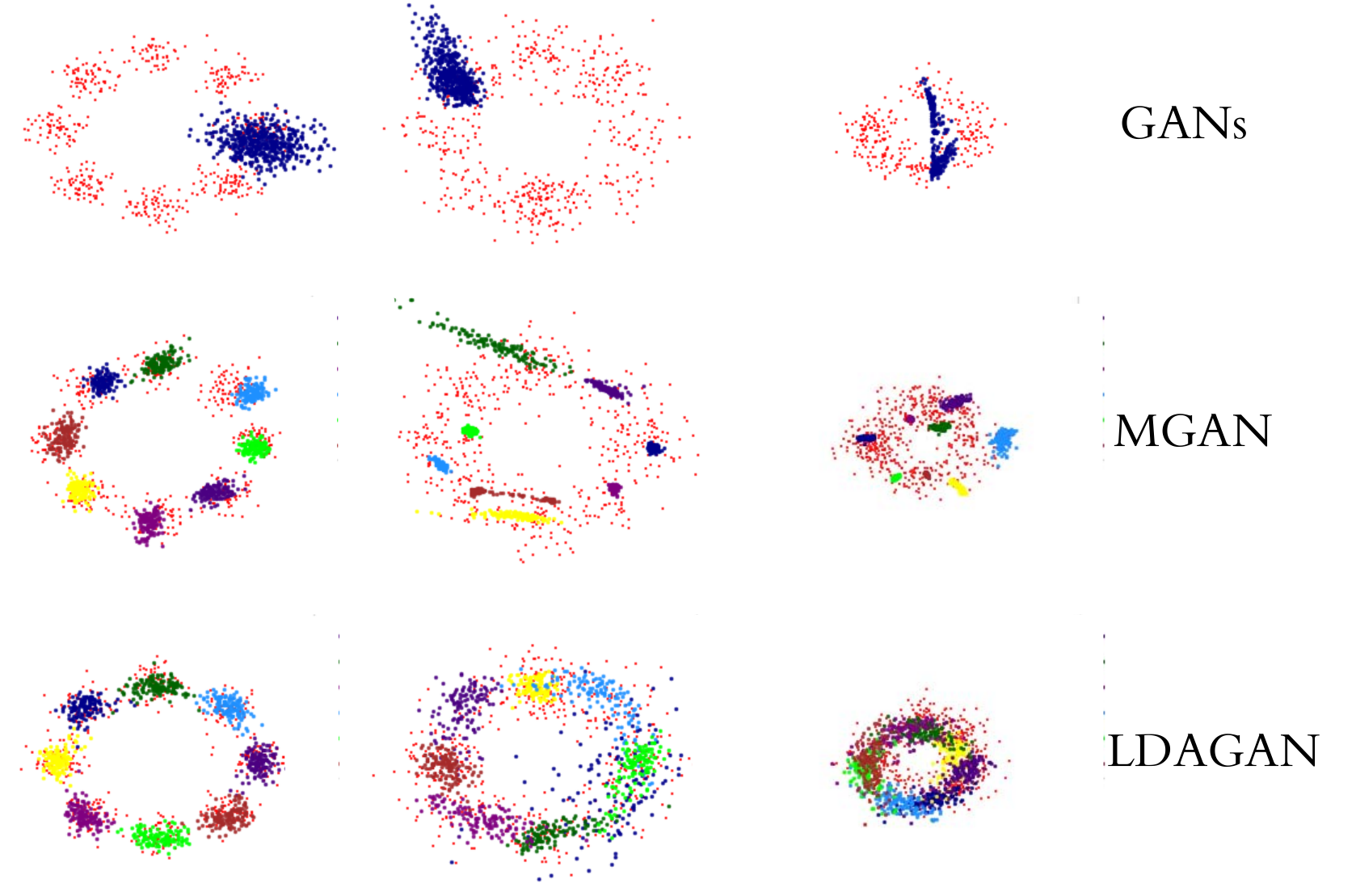}
     \caption{The generative performance comparison of LDAGAN and other GANs on synthetic data. Red points denote real data and blue points denote data generated by GANs. In MGAN and LDAGAN, points in different color mean data generated by different generators.}
     \label{fig:Synth}
\end{figure}

We first evaluated the performance of LDAGAN on a synthetic dataset with 3 types of data.
The fist type of training data was sampled from a 2D mixture of 8 isotopic Gaussians, where the mixing weights are fixed to be 0.125.
The means of all Gaussians uniformly distribute on a circle, the center and radius of which are $\mathbf{0}$ and 2.0, and the covariance matrices of all Gaussians are all $0.08\mathbf{I}$.
The second type of training data was drawn from the distribution of a LDA generative model, where $\bm{\pi}$ had a Dirichlet prior with $\bm{\alpha} = [8,4,...,8,4]$.
The third type of training data is similar to the second, but consists of different Gaussians.
The means of all Gaussians distribute on a circle with a much smaller radius $0.5$ and the covariance matrices of all Gaussians are $0.02\mathbf{I}$.
Each type of training data is shown in different columns of Fig.~\ref{fig:Synth}.

On this dataset we compared three types of GANs: (\romannumeral1)  GANs, (\romannumeral2)  MGAN, and (\romannumeral3) LDAGAN.
For MGAN, we employed 8 generators, each of which was designed to have an input layer with 256 units and two fully connected hidden layers with 128 ReLU units.
The network architectures of discriminator and classifier in MGAN were constructed following~\cite{hoang2018mgan}.
For LDAGAN, we also employed 8 generators and each had the same input layer as MGAN, but only one fully connected hidden layer with 128 ReLU units.
The Dirichlet parameters $\bm{\alpha}$ were all initialized to be 2.

\begin{figure*}[t]
  \centering
    \subfigure[CIFAR-10]{\includegraphics[width=0.3\textwidth]{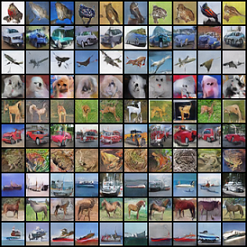}}
    \hspace{.15in}
    \subfigure[CIFAR-100]{\includegraphics[width=0.3\textwidth]{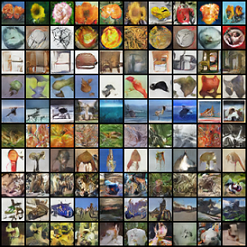}}
    \hspace{.15in}
    \subfigure[ImageNet]{\includegraphics[width=0.3\textwidth]{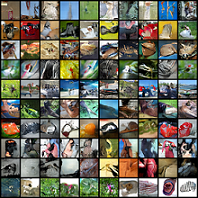}}
     \caption{Images (with size $32\times32$) generated by different generators of LDAGAN. Each row corresponds to one generator (\ie~mode). \textbf{(a):} Trained on CIFAR-10. The images generated by the same generator have highly similarity, for example, the ``car'' images in the 2rd row, the ``dog'' images in the 4th row, and the ``ship'' images in the last row.  \textbf{(b):} Trained on CIFAR-100. Obvious image similarity can be found in the same row, such as the 2rd and 7th rows. \textbf{(c):} Trained on ImageNet.}
     \label{fig:Struc}
\end{figure*}

Visible results of these experiments can be found in Fig.~\ref{fig:Synth}.
It shows the fitting results of 512 samples generated by LDAGAN and other baseline methods.
A consistent trend behind this is that LDAGAN captures data modes more precisely than MGAN and other GANs.
As discussed in Sec.~\ref{sec:Previous} and Sec.~\ref{sec:LDAGAN}, this is due to its ability to learn underlying structure of real data and optimize generators based on the learned structure.
Single generator based GANs fails to account for capturing all modes of data, leading to mode collapse.
Because MGAN simply mix generators together and encourage mode diversity of synthetic samples, incurring intra-class mode dropping problem.

\subsection{Evaluation on Real-Word Data}

Testing the performance of LDAGAN on real-word data is more meaningful as it gives a better indication on how well the method generalizes.
To further evaluate the effectiveness of LDAGAN, we tested it on large-scale real-word datasets.

\subsubsection{Datasets and Evaluation Metrics.}
We used 3 challenging real-word datasets to demonstrate the effectiveness of our proposed LDAGAN.
The details of the three datasets are described in the following.

\noindent\textbf{CIFAR-10}: It has 60,000 labeled $32\times32$-sized RGB natural images in 10 classless.
The 10 classes include: airplane, automobile, bird, cat, deer, dog, frog, horse, ship and truck.
There are 50000 training images.

\noindent\textbf{CIFAR-100}: It is just like the CIFAR-10, but it has much more diverse classes.
It has 100 classes containing 600 images each.
There are 50000 training images.

\noindent\textbf{ImageNet}:  It contains over 1.4 million images of 1000 classes.
It is the largest,  most diverse, and most significant visual dataset at present.

To conduct fair comparison with the baselines, we resized images in ImageNet dataset to $32\times32$.
On the above three datasets, we used inception score and Fr\'{e}chet inception distance for performance evaluation.
\subsubsection{Model Architecture.}
Our generator and discriminator architectures follow the design of DCGANs and ResNet based
GANs.
Moreover, all generators share parameters except the first layer.
This parameter sharing scheme helps to balance the learning of generators since we have only one discriminator.
In DCGANs architecture, considering the declining problem of active neurons, we fixed the batch normalization center to be zero for all layers in the generator networks as in~\cite{hoang2018mgan}.
In ResNet based GANs architecture, we did not fixed the batch normalization center.
Please see Appendix B for the details.

\subsubsection{Parameter and Hyperparameter Settings.}
The hyperparameters of LDAGANs includes: the number of generators $K$ and the minibatch size.
For standard CNN based architecture, the number of generators was set to be 10 on CIFAR-10 and CIFAR-100, and 20 on ImageNet.
We also trained models with 10, 27, 30 generators on ImageNet, and observed the best performance was achieved by the model with 20 generators (see  Appendix D).
We used a minibatch size of 24 for each generator on CIFAR-10 and CIFAR-100, and 12 on ImageNet.
For ResNet based architecture, the number of generators was all set to be 10 on the above three datasets.
We used a minibatch size of 12 for each generator.
The Dirichlet parameters $\bm{\alpha}$ were initialized to be 8.
Due to the importance of parameter sharing scheme in LDAGAN, we investigated how sharing scheme impacted its performance.
The best inception score and FID were exhibited when we removed the parameter sharing in the first hidden layer (see Appendix C).

\subsubsection{Underlying Mode Finding.}
Example images generated by LDAGAN can be found in Fig.~\ref{fig:Struc}.
The results show a consistent phenomenon that images in one row, which are generated by the same generator, have highly similarity.
For exapmle, in Fig.~\ref{fig:Struc}a, we observe the ``car'' images in the second row, the ``dog'' images in the fourth row, and the ``ship'' images in the last row.
As discussed in Sec.~\ref{sec:LDAGAN}, this is due to LDAGAN's capability to find the underlying structure (\ie~mode) of real data and guide multiple generators to fit the structured data.
The similar phenomenon can also be found in both Fig.~\ref{fig:Struc}b and Fig.~\ref{fig:Struc}c.
\begin{table*}[t]
\caption{Inception Score(the higher the better) and Fr\'{e}chet Inception Distance(the lower the better) on different datasets. $*$ denotes the results obtained by running the released codes of authors.}
\label{Tab:FID}
\centering
\addtolength{\tabcolsep}{-0pt}
\resizebox{1.0\textwidth}{!}{
{\begin{tabular}{lccc|cccc}
\hline

\hline
                &\multicolumn{3}{c}{\textbf{IS}}      & \multicolumn{4}{c}{\textbf{FID}}  \tabularnewline
                & CIFAR-10 & CIFAR-100 & ImageNet  & \multicolumn{2}{c}{CIFAR-10}  & CIFAR-100    & ImageNet\tabularnewline
\hline
\hline
\textbf{-Standard CNN-}\tabularnewline
DCGAN~\cite{radford2015unsupervised} &  $6.40^{~}$        &    $6.97^*$   &  $7.89^{~}$ &      $37.7^{~}$&-     &   $36.1^*$    &   $37.4^*$  \tabularnewline

WGAN-GP~\cite{gulrajani2017improved} &  $6.68^{~}$        &   $6.79^*$    & $7.58^*$  &      $29.3^{~}$&-    &    $35.1^*$    & $33.2^*$  \tabularnewline

WGAN-GP+TTUR~\cite{heusel2017gans}  &-&-& - &      $24.8^{~}$&-    &         -       &    -    \tabularnewline

SNGAN~\cite{miyato2018spectral}       &   $7.42^{~}$       &         -             &    -  &      -&$29.3^{~}$  &         -       &    -     \tabularnewline

MGAN~\cite{hoang2018mgan} &   $7.30^*$         &    $\mathbf{7.67}^*$     &   $7.59^*$ &$26.7^{~}$  &-     &     $32.9^*$ &  $36.8^*$\tabularnewline

Graphical GAN~\cite{NIPS2018_7846} &   $5.94^{~}$       &   $5.64^*$   &   $6.43^*$   &      $55.8^*$&-     &       $55.9^*$    &   $48.8^*$   \tabularnewline

LDAGAN         &$\mathbf{7.46}^{~}$ &       $7.50^{~}$           &   $\mathbf{8.37}$      & $\mathbf{24.3}^{~}$&$\mathbf{28.3}$    & $\mathbf{28.8}^{~}$ & $\mathbf{28.9}^{~}$ \tabularnewline
 \hline
\textbf{ -ResNet- }\tabularnewline
 MGAN-SN~\cite{hoang2018mgan}   &$8.18^*$         &       $8.53^*$       & $8.52^*$     &     $12.7^*$&-       &   $15.5^*$      &    $23.3^*$   \tabularnewline
 LDAGAN-SN  &$\mathbf{8.77}^{~}$&$\mathbf{8.81}^{~}$&$\mathbf{9.70}^{~}$     &$\mathbf{10.4}^{~}$&-   &$\mathbf{15.2}^{~}$&$\mathbf{18.5}^{~}$\tabularnewline
 \hline
\end{tabular}}
}
\end{table*}
\begin{figure*}[t]
\centering
\includegraphics[width=0.8\textwidth]{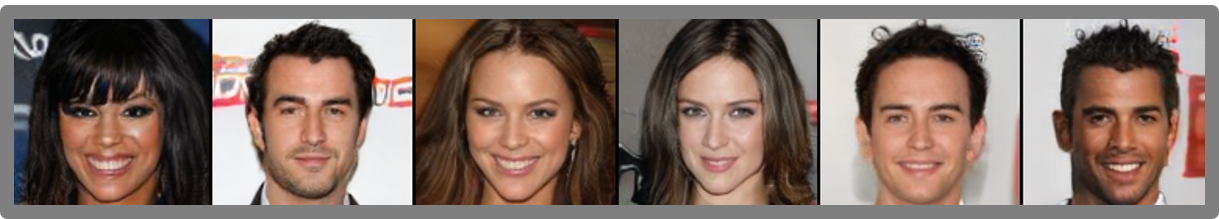}
     \caption{LDAGAN samples on CelebA $128\times128$.}
     \label{fig:HRImg}
\end{figure*}

\subsubsection{Image Quality Comparison.}
\label{sec:Comp}
On the real-word datasets, we compared LDAGAN with some other state-of-the-art GANs.
For MGAN~\cite{hoang2018mgan}, we set the number of generators to be 10 and set all the other hyper-parameters following.
For SNGANs~\cite{miyato2018spectral}, we reported two versions of FID calculated using 50000 and 5000 generated images to facilitate comparison.
For Graphical GAN~\cite{NIPS2018_7846}, we followed the original work GMGAN~\cite{NIPS2018_7846} and fixed the number of components
in the mixture model to be 30 and 50 on CIFAR-10 and CIFAR-100 datasets, respectively.
For LDAGAN, we set the parameters and hyper-parameters as described in previous experimental setting. 

The results in Tab.~\ref{Tab:FID} show significant tendency in the comparison with Graphical-GAN, MGAN and other single-generator based GANs:
(\romannumeral1) LDAGAN significantly outperforms Graphical-GAN as its inception score increases by 1.52 and 1.86, and its FID decreases by 33.1 and 27.1 on the CIFAR-10 and CIFAR-100 datasets.
As discussed in Sec.~\ref{sec:LDAGAN}, this is due to LDAGAN's ability to precisely model the multimodal generation process of images.
However, Graphical-GAN does not define the proper one.
(\romannumeral2) MGAN\footnote[1]{For MGAN, we ran the codes released by the author, but had not obtain the IS 8.33 and 9.32 on CIFAR-10 and ImageNet. Thus, we reported the best results we obtained by running the released codes in Tab.~\ref{Tab:FID}.} shows an improved IS and FID over Graphical-GAN.
However, it also has inferior performance to LDAGAN because of its simple mixing strategies, which can not account for data structure.
This can be seen by its poorer FID, for example 36.8 on the ImageNet dataset with CNN architecture.
(\romannumeral3)  LDAGAN exhibits better performance than most existing single-generator based GANs, such as WGAN-GP+TURR and SNGAN, which have shown state-of-the-art results.

\subsubsection{High Resolution Image Generation.}
On the CelebA dataset~\cite{liu2015faceattributes}, we generated images of size $128\times128$.
Example images generated by LDAGAN can be found in Fig.~\ref{fig:HRImg}.
More details are described in Appendix H.

\section{Conclusion}
\label{sec:Conclusion}
Latent Dirichlet allocation was introduced into generative adversarial networks in this work.
It helps to discover the multimodal generation mechanism of vision data, and make generators better fit data.
Moreover, EM algorithm was combined with adversarial technique to solve our model.
The proposed method was shown to outperform the existing GANs with different network architecture for IS and FID score.
Our future work will involve extending LDAGAN to more computer vision problems.

\bibliography{example_paper}
\bibliographystyle{icml2018}

\onecolumn
\appendix
\section{Variational EM algorithm}
\label{Appendix}

In this section, we derive the variational EM algorithm for efficient inference in the adversarial training,  described in Sec.~4.
\subsection{Variational Distribution}
\label{App:VaDistribution}

Based on mean-field approximation, we define a variational distribution which factorizes between latent variables $\bm{\pi}$ and $\bm{z}$ so that:
\begin{equation}
\label{eq:AVarDis1}
q\left( \bm{\pi}, \mathbf{z}|\bm{\gamma},\bm{\omega} \right)
=
q\left( \bm{\pi} | \bm{\gamma} \right)
q\left( \mathbf{z} | \bm{\omega} \right),
\end{equation}
where $\bm{\gamma}$ and $\bm{\omega}$ are the Dirichlet and multinomial parameters respectively.
It can be viewed as a surrogate for the posterior $p\left(\bm{\pi}, \mathbf{z}, y=1|\mathbf{z}',\bm{\theta}, \bm{\alpha},\bm{\phi}\right)$.

The log likelihood of a sample generated from $\mathbf{z}'$ being real is obtained by summing the joint distribution over all possible modes $\mathbf{z}$ and integrating over all mode distributions $\bm{\pi}$:
\begin{equation}
\label{eq:AVarDis2}
\log p\left( y=1 | \mathbf{z}',\bm{\theta},\bm{\alpha}, \bm{\phi} \right)
=
\log
\int
\sum_{\mathbf{z}}
p\left( \bm{\pi},\mathbf{z},y=1 | \mathbf{z}',\bm{\theta},\bm{\alpha},\bm{\phi} \right)
d\bm{\pi}.
\end{equation}
\noindent According to Jensen's inequality, the above log likelihood function has a lower bound:
\begin{align}
\label{eq:AVarDis3}
\log
\int
\sum_{\mathbf{z}}
\frac{p\left( \bm{\pi},\mathbf{z},y=1 | \mathbf{z}',\bm{\theta},\bm{\alpha},\bm{\phi} \right)
q\left( \bm{\pi},\mathbf{z} | \bm{\gamma},\bm{\omega} \right)}
{q\left( \bm{\pi}, \mathbf{z} | \bm{\gamma},\bm{\omega} \right)}
d\bm{\pi}
&\geq  \nonumber
\int
\sum_{\mathbf{z}}
q\left( \bm{\pi}, \mathbf{z}|\bm{\gamma},\bm{\omega} \right)
\log p\left( \bm{\pi},\mathbf{z},y=1 | \mathbf{z}',\bm{\theta},\bm{\alpha},\bm{\phi} \right)
d\bm{\pi}  \\
&-\int
\sum_{\mathbf{z}}
q\left( \bm{\pi},\mathbf{z} | \bm{\gamma},\bm{\omega} \right)
\log q\left( \bm{\pi},\mathbf{z} | \bm{\gamma},\bm{\omega} \right)
d\bm{\pi},
\end{align}
which is denoted as the right-hand side of Eq.~(\ref{eq:AVarDis3}), represented by $L\left(\bm{\gamma},\bm{\omega};\bm{\theta},\bm{\alpha}\right)$.
Thus, the log likelihood has the following decomposition:
\begin{equation}
\label{eq:AVarDis4}
\log p\left(y=1|\mathbf{z}',\bm{\theta},\bm{\alpha}, \bm{\phi}\right)
=
L\left( \bm{\gamma},\bm{\omega};\bm{\alpha},\bm{\theta} \right)
+ \text{KL}\left( q\left( \bm{\pi},\mathbf{z} | \bm{\gamma},\bm{\omega} \right)
|| p\left( \bm{\pi}, \mathbf{z} |y=1, \mathbf{z}',\bm{\theta},\bm{\alpha},\bm{\phi} \right) \right),
\end{equation}
which is the sum of the lower bound and the Kullback-Leibler divergence.
\subsection{E-Step Optimization}
In E-step, we maximize the lower bound $L\left(\bm{\gamma},\bm{\omega};\bm{\theta},\bm{\alpha}\right)$ with respect to variational parameters $\bm{\gamma}$ and $\bm{\omega}$.
It can be easily verified that the maximum of $L\left(\bm{\gamma},\bm{\omega};\bm{\theta},\bm{\alpha}\right)$ occurs when the KL divergence vanishes.
We expand the lower bound by using the factorizations of $p\left(\bm{\pi},\mathbf{z},y=1|\mathbf{z}',\bm{\theta},\bm{\alpha},\bm{\phi}\right)$ and $q\left(\bm{\pi}, \mathbf{z}|\bm{\gamma},\bm{\omega}\right)$:
\begin{equation}
\label{eq:AVarDis5}
L\left( \bm{\gamma},\bm{\omega}; \bm{\alpha},\bm{\theta} \right)
 =
\mathbb{E}_q \left[ \log  p\left( \bm{\pi} | \bm{\alpha} \right) \right]
+\mathbb{E}_q\left[ \log p\left( \mathbf{z} | \bm{\pi} \right) \right]
+\mathbb{E}_q\left[ \log p\left( y=1 | \mathbf{z},\mathbf{z}',\bm{\theta},\bm{\phi}\right) \right]
-\mathbb{E}_q\left[ \log  q\left( \bm{\pi} | \bm{\gamma} \right) \right]
-\mathbb{E}_q\left[ \log  q\left( \mathbf{z} | \bm{\omega} \right) \right],
\end{equation}
where $\mathbb{E}_q\left[\cdot\right]$ denotes the expectation with respect to variational distribution.
Each of the terms in the sum in Eq.~(\ref{eq:AVarDis5}) has the following expressions.
The first term is:
\begin{equation}
\label{eq:AVarDis6}
\mathbb{E}_q \left[ \log  p\left( \bm{\pi} | \bm{\alpha} \right) \right]
=
\log\Gamma\left( \sum\nolimits_{j=1}^K \alpha_j \right)
- \sum_{k=1}^K \log\Gamma\left(\alpha_k\right)
+\sum_{k=1}^K
\left( \alpha_k-1 \right)
\left( \Psi\left(\gamma_k\right)-\Psi\left(\sum\nolimits_{j=1}^K\gamma_j\right) \right),
\end{equation}
where $\Psi$ is the digamma function, representing the first derivative of the log Gamma function.
Similar, the remaining terms of Eq.~(\ref{eq:AVarDis5}) have the forms of:
\begin{align}
\mathbb{E}_q\left[ \log p\left( \mathbf{z} | \bm{\pi} \right) \right]
=&
\sum_{k=1}^K
\omega_k
\left( \Psi\left( \gamma_k \right) - \Psi\left( \sum\nolimits_{j=1}^K \gamma_j \right)\right),\label{eq:AVarDis7}\\
\mathbb{E}_q\left[ \log p\left( y=1 | \mathbf{z},\mathbf{z}',\bm{\theta},\bm{\phi}\right) \right]
=&
\sum_{k=1}^K
\omega_k
\log D\left( G_k\left( \mathbf{z}'\right) \right),\label{eq:AVarDis8}\\
\mathbb{E}_q \left[ \log  q\left( \bm{\pi} | \bm{\gamma} \right) \right]
=&
\log\Gamma\left( \sum\nolimits_{j=1}^K \gamma_j \right)
- \sum_{k=1}^K \log\Gamma\left(\gamma_k\right)\nonumber\\
&+\sum_{k=1}^K
\left( \gamma_k-1 \right)
\left( \Psi\left(\gamma_k\right)-\Psi\left(\sum\nolimits_{j=1}^K\gamma_j\right) \right),\label{eq:AVarDis9}\\
\mathbb{E}_q\left[ \log  q\left( \mathbf{z} | \bm{\omega} \right) \right]
=&
\sum_{k=1}^K
\omega_k
\log \omega_k.
\label{eq:AVarDis10}
\end{align}
Here, we use the fact that the expected value of the log of a single component under the Dirichlet have the following expression $\mathbb{E}_q\left[\log \pi_k\right] = \Psi\left(\bm{\gamma}_k\right)-\Psi\left(\sum_{j=1}^K \bm{\gamma}_j\right)$~\cite{blei2003latent}.

In the next two sections, we will show how to maximize the lower bound with respect to the variational Dirichlet and multinomial parameters $\bm{\gamma}$ and $\bm{\omega}$.
\subsubsection{Variational Dirichlet}
\label{App:VaDirichlet}

Picking out just those terms that only contain $\bm{\gamma}$ in $L\left(\bm{\gamma},\bm{\omega};\bm{\theta},\bm{\alpha}\right)$, we have:
\begin{align}
\label{eq:Appendix15}
L_{\bm{\gamma}}
&=
\sum_{k=1}^K
\left( \alpha_k-1 \right)
\left( \Psi\left(\gamma_k\right)-\Psi\left(\sum\nolimits_{j=1}^K\gamma_j\right) \right)
+\sum_{k=1}^K
\omega_k
\left(\Psi \left( \gamma_k \right)-\Psi \left( \sum\nolimits_{j=1}^K \gamma_j \right) \right) \nonumber \\
&-\log \Gamma \left( \sum\nolimits_{j=1}^K\gamma_j\right)
+\sum_{k=1}^K\log \Gamma \left( \gamma_k \right)
-\sum\nolimits_{k=1}^K
\left( \gamma_k-1 \right)
\left( \Psi\left( \gamma_k \right) -\Psi \left( \sum\nolimits_{j=1}^K \gamma_j\right) \right),
\end{align}
\noindent which simplifies to:
\begin{equation}
\label{eq:Appendix16}
L_{\bm{\gamma}}
=
\sum_{k=1}^K
\left( \Psi\left( \gamma_k \right) -\Psi \left( \sum\nolimits_{j=1}^K \gamma_j\right) \right)
\left( \alpha_k + \omega_k - \gamma_k \right)
-\log \Gamma \left( \sum\nolimits_{j=1}^K\gamma_j\right)
+\sum_{k=1}^K\log \Gamma \left( \gamma_k \right).
\end{equation}
\noindent The derivative of $L_{\bm{\gamma}}$ with respect to $\gamma_k$ can be expressed as:
\begin{equation}
\label{eq:Appendix17}
\frac{\partial L_{\bm{\gamma}}}
{\partial{\gamma_k}}
=
\Psi'\left( \gamma_k \right)
\left( \alpha_k + \omega_k - \gamma_k \right)
- \Psi'\left( \sum\nolimits_{j=1}^K \gamma_j \right)
\sum_{j=1}^K \left(\alpha_j + \omega_j - \gamma_j \right),
\end{equation}
where $\gamma_k$ is the $k^{th}$ element of $\bm{\gamma}$. Setting this derivative to zero yields a maximum at:
\begin{equation}
\label{eq:Appendix18}
\gamma_k
=
\alpha_k
+
\omega_k.
\end{equation}

\subsubsection{Variational Multinomial}
\label{App:VaMultinomial}

In this section, we show how to maximize the lower bound $L\left( \bm{\gamma},\bm{\omega}; \bm{\theta},\bm{\alpha} \right)$ with respect to the variational parameters $\bm{\omega}$.
One note there is a constrain on $\bm{\omega}$, which is $\sum_{k=1}^K\omega_k=1$.
We form the Lagrange by isolating the terms with respect to $\omega_k$, where $\omega_k$ is the $k^{th}$ element of $\bm{\omega}$, and adding the Lagrange multipliers,
\begin{equation}
\label{eq:Append.1.1-1}
L_{\omega_k}
=
\omega_k\left( \Psi \left( \gamma_k \right) - \Psi \left( \sum\nolimits_{j=1}^K \gamma_j \right) \right)
+ \omega_k \log D\left( G_k\left( \mathbf{z}' \right) \right)
- \omega_k \log \omega_k
+\lambda \left( \sum\nolimits_{j=1}^K \omega_j - 1\right).
\end{equation}
\noindent The derivatives of $L_{\omega_k}$ with respect to $\omega_k$ is given by:
\begin{equation}
\label{eq:Appendix13}
\frac{\partial L_{\omega_k}}{\partial \omega_k}
=
\Psi \left( \gamma_k \right) - \Psi \left( \sum\nolimits_{j=1}^K \gamma_j \right) + \log D\left( G_k\left( \mathbf{z}'\right) \right)
- \log \omega_k
- 1
+\lambda.
\end{equation}
Setting this derivative to zero yields the optimal value of variational parameter $\omega_k$:
\begin{equation}
\label{eq:Appendix14}
\omega_k
\propto
D\left( G_k\left( \mathbf{z}'\right) \right)\exp\left( \Psi\left(\gamma_k\right) - \Psi \left( \sum\nolimits_{j=1}^K \gamma_j\right) \right).
\end{equation}

\subsection{M-Step Optimization}
\label{App:PaEstimation}

In the subsequent M-step, the variational distribution $q\left( \bm{\pi}, \mathbf{z}|\bm{\gamma},\bm{\omega} \right)$ is fixed and the lower bound is maximized with respect to model parameters $\bm{\theta}$ and $\bm{\alpha}$.
In previous discussion, one note that we only consider the log likelihood for a single sample $\mathbf{z}'$ and the estimate of $\bm{\gamma}$ and $\bm{\omega}$ (see Eq.~(\ref{eq:Appendix14}) and Eq.~(\ref{eq:Appendix18})) are related to $\mathbf{z}'$.
We thus rewrite $\bm{\gamma}$ and $\bm{\omega}$ as functions of $\mathbf{z}'$, denoted by $\bm{\gamma}\left(\mathbf{z}'\right)$ and $\bm{\omega}\left(\mathbf{z}'\right)$.
Estimating model parameters $\bm{\theta}$ and $\bm{\alpha}$ should consider the lower bounds over all possible $\mathbf{z}'$, that is, the expectation of $L\left( \bm{\gamma},\bm{\omega}; \bm{\alpha},\bm{\theta} \right)$ with respect to $\mathbf{z}'$, described by $\mathcal{L}\left( \bm{\gamma},\bm{\omega}; \bm{\alpha},\bm{\theta}\right) = \mathbb{E}_{\mathbf{z}'\sim p\left(\mathbf{z}'\right)}\left[L\left( \bm{\gamma},\bm{\omega}; \bm{\alpha},\bm{\theta}\right)\right]$.

\subsubsection{Generators}
\label{App:Generators}

To optimize $\bm{\theta}_k$ associated with the $k^{th}$ generator, we isolate the terms in $\mathcal{L}\left( \bm{\gamma},\bm{\omega}; \bm{\alpha},\bm{\theta}\right)$ containing $\bm{\theta}_k$ and obtain:
\begin{equation}
\label{eq:Appendix19}
\mathcal{L}_{\bm{\theta}_k}
=
\mathbb{E}_{\mathbf{z}' \sim p\left(\mathbf{z}'\right)}
\left[
\omega_k\left( \mathbf{z}' \right)
\log D\left( G_k\left( \mathbf{z}'\right) \right)
\right],
\end{equation}
maximizing $\mathcal{L}_{\bm{\theta}_k}$ yields the following optimization problem:
\begin{equation}
\label{eq:Appendix20}
\max_{\bm{\theta}_k}
\mathbb{E}_{\mathbf{z}' \sim p \left( \mathbf{z}' \right)}
\left[
\omega_k\left( \mathbf{z}' \right)
\log D\left( G_k\left( \mathbf{z}' \right) \right)
\right].
\end{equation}
\subsubsection{Dirichlet}
\label{App:Dirichlet}

The terms in $\mathcal{L}\left( \bm{\gamma},\bm{\omega}; \bm{\alpha},\bm{\theta}\right)$ which contain $\bm{\alpha}$ are:
\begin{equation}
\label{eq:Appendix21}
\mathcal{L}_{\bm{\alpha}}
=
\log \Gamma \left( \sum\nolimits_{j=1}^K \alpha_j \right)
-\sum_{k=1}^K \log \Gamma \left( \alpha_k \right)
+
\mathbb{E}_{\mathbf{z}' \sim p_{\mathbf{z}'}}
\left[
\sum_{k=1}^K
\left( \alpha_k-1 \right)
\left( \Psi\left(\gamma_k\left(\mathbf{z}'\right)\right)
-\Psi\left(\sum\nolimits_{j=1}^K\gamma_j\left(\mathbf{z}'\right)\right) \right)
\right].
\end{equation}
Taking the derivative of $\mathcal{L}_{\bm{\alpha}}$ with respect to $\alpha_k$, we have:
\begin{equation}
\label{eq:Appendix22}
\frac{\partial \mathcal{L}_{\bm{\alpha}}}
{\partial \alpha_k}
=
\Psi\left(\sum\nolimits_{j=1}^K\alpha_j\right)-\Psi\left(\alpha_k\right)
+
\mathbb{E}_{\mathbf{z}' \sim p_{\mathbf{z}'}}
\left[
\Psi\left( \gamma_k\left( \mathbf{z}' \right) \right)
-\Psi\left( \sum\nolimits_{j=1}^K \gamma_j\left( \mathbf{z}' \right) \right)
\right].
\end{equation}
Finally, we use gradient ascent to update $\bm{\theta}$ and $\bm{\alpha}$.

\section{Network Architecture}
\label{App:NetArch}

\begin{figure}[ht]
  \centering
    \includegraphics[width=0.7\textwidth]{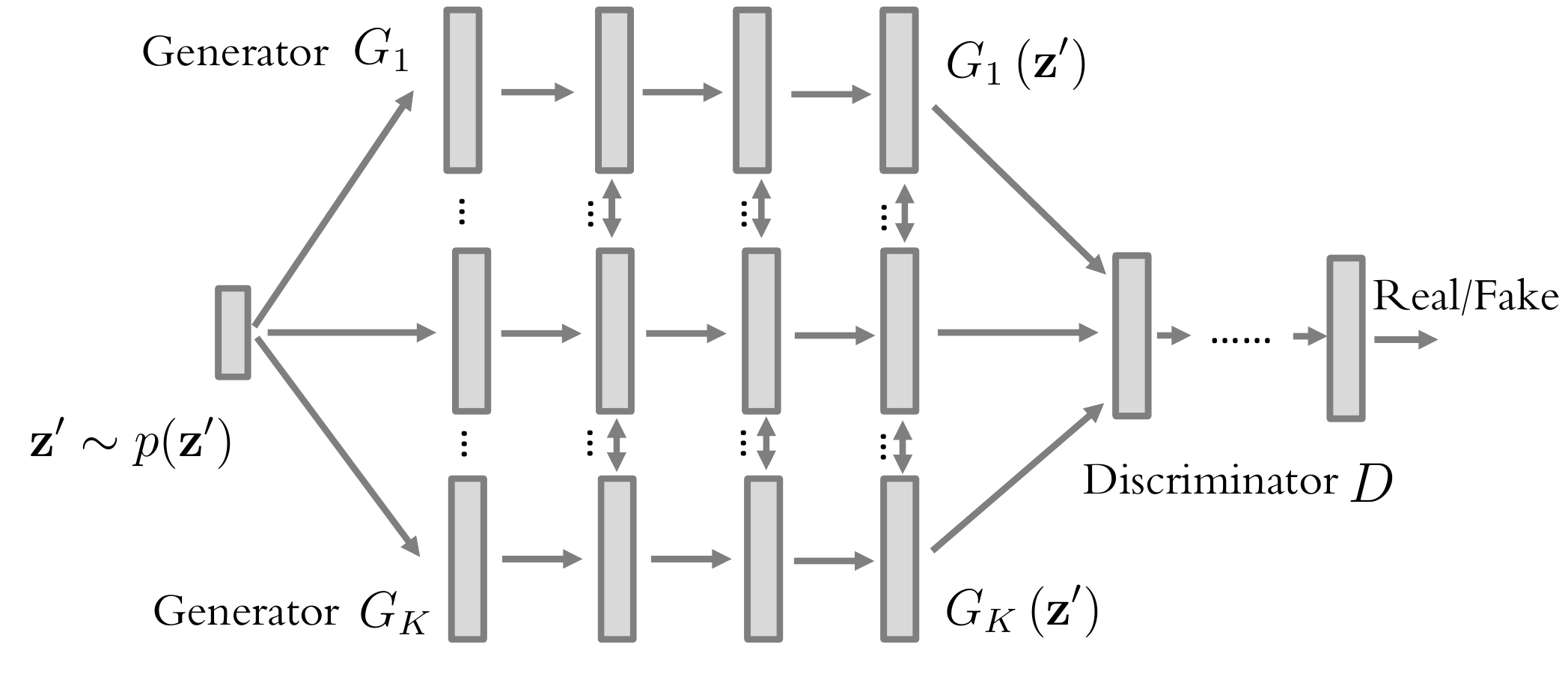}
     \caption{Network architecture of LDAGAN (standard CNN based), which has $K$ generators and 1 discriminator. Each generator has four convolutional layers, and parameter sharing occurs on the last three layers.}
     \label{fig:Arch}
\end{figure}

Our LDAGAN consists of $K$ generators and 1 discriminator, as illustrated in Fig.~\ref{fig:Arch}.
Each generator has four convolutional layers, and shares parameters except the input layer.
The parameter sharing scheme helps to keep the balance of generator's updating.
Besides, it dramatically reduces the number of parameters, and thus ensures the training process, compared with standard GANs, is not time consuming.

We constructed the CNN based network for LDAGAN according to the design of DCGAN~\cite{radford2015unsupervised} with some slight modifications.
The details our networks trained on the CIFAR-10, CIFAR-100, and ImageNet datasets can be found in Tab.~\ref{tab1} and Tab.~\ref{tab2}.
``BN'' is short for batch normalization, and ``Shared'' is the short for parameter sharing.
Moreover, we constructed the ResNet based network for LDAGAN according to ResNet based GANs.

\begin{table}[H]
\small
\centering\caption{The CNN based Network Architecture and hyperparameters of LDAGAN (CIFAR-10 and CIFAR-100).}
\resizebox{.95\textwidth}{!}{
\centering
{
\begin{tabular}{r l c c r c l c}
\hline
Unit & Operation & Kernel & Stride & Feature maps & BN & Activation & Shared\\ 
\hline
\hline
Generator& $\mathbf{z}'$ $\sim$ Uniform[-1,1] & & & 100 & & &\\
& Conv transposed & 4x4 & 1 & 128x4 & Yes & ReLU & No\\
& Conv transposed & 4x4 & 2 & 128x2 & Yes & ReLU & Yes\\
& Conv transposed & 4x4 & 2 & 128 & Yes & ReLU & Yes\\
& Conv transposed & 4x4 & 2 & 3 & No & Tanh & Yes\\
\hline
Discriminator & Conv & 5x5 & 2 & 128 & No & LReLU &\\
& Conv  & 5x5 & 2 & 128x2 & Yes & LReLU &\\
& Conv  & 5x5 & 2 & 128x4 & Yes & LReLU &\\
& Conv  & 4x4 & 1 & 1 & Yes & Sigmoid &\\
\hline
Number of generators& 10& & & & & &\\
Generator initialization  &$\mathcal{N}\left(\mu=0,\sigma=0.08\right)$ & & & & & &\\
Discriminator initialization  &$\mathcal{N}\left(\mu=0,\sigma=0.02\right)$ & & & & & &\\
Batch size of real data  &64 & & & & & &\\
Batch size for each generator  &12 & & & & & &\\
Leacky ReLU slope  &0.2 & & & & & &\\
Learning rate  &0.0001 & & & & & &\\
Optimizer  &Adam($0.5$, $0.999$) & & & & & &\\
\hline
\end{tabular}}}
\label{tab1}
\end{table}

\begin{table}[H]
\small
\centering\caption{The CNN based Network Architecture and hyperparameters of LDAGAN (ImageNet).}
\resizebox{.95\textwidth}{!}{
\centering
{
\begin{tabular}{r l c c r c l c}
\hline
Unit & Operation & Kernel & Stride & Feature maps & BN & Activation & Shared\\ 
\hline
\hline
Generator& $\mathbf{z}'$ $\sim$ Uniform[-1,1] & & & 100 & & &\\
& Conv transposed & 4x4 & 1 & 128x4 & Yes & ReLU & No\\
& Conv transposed & 5x5 & 2 & 128x2 & Yes & ReLU & Yes\\
& Conv transposed & 5x5 & 2 & 128 & Yes & ReLU & Yes\\
& Conv transposed & 5x5 & 2 & 3 & No & Tanh & Yes\\
\hline
Discriminator & Conv & 5x5 & 2 & 128 & No & LReLU &\\
& Conv  & 5x5 & 2 & 128x2 & Yes & LReLU &\\
& Conv  & 5x5 & 2 & 128x4 & Yes & LReLU &\\
& Conv  & 4x4 & 1 & 1 & Yes & Sigmoid &\\
\hline
Number of generators& 20& & & & & &\\
Generator initialization  &$\mathcal{N}\left(\mu=0,\sigma=0.08\right)$ & & & & & &\\
Discriminator initialization  &$\mathcal{N}\left(\mu=0,\sigma=0.02\right)$ & & & & & &\\
Batch size of real data  &64 & & & & & &\\
Batch size for each generator  &24 & & & & & &\\
Leacky ReLU slope  &0.2 & & & & & &\\
Learning rate  &0.0001 & & & & & &\\
Optimizer  &Adam($0.5$, $0.999$) & & & & & &\\
\hline
\end{tabular}}}
\label{tab2}
\end{table}
\begin{table}[H]
\small
\centering\caption{The ResNet based Network Architecture and Hyperparameters of LDAGAN (CIFAR10, CIFAR100 and ImageNet).}
\centering
{
\begin{tabular}{r l c c r c}
\hline
Unit & Operation & Input kernels & Output kernels  & Shared\\ 
\hline
\hline
Generator& $\mathbf{z}'$ $\sim$ Normal(0, 1) & & & \\
& Linear & 128 & 256x4x4 & No\\
& GenResBlock & 256 & 256 & Yes\\
& GenResBlock & 256 & 256 & Yes\\
& GenResBlock & 256 & 256 & Yes\\
& Conv(Tanh) & 256 & 3 & Yes\\
\hline
Discriminator & OptimizeResblock & 3 & 128 &\\
& DisResBlock & 128 & 128 & \\
& DisResBlock & 128 & 128 &\\
& DisResBlock & 128 & 128 &\\
& Global Average Pooling & & &\\
& Linear & 128 & 1 &\\
\hline
Number of generators& 10& & & \\
Generator initialization
&Xavier\_Uniform $\left( \sqrt{2} \right)$ & & &\\
Discriminator initialization
&Xavier\_Uniform $\left( \sqrt{2} \right)$ & & &\\
Batch size of real data  &64 & & & \\
Batch size for each generator  &12 & & &\\
Learning rate  &0.0002 & & &\\
Optimizer  &Adam($0$, $0.9$) & & &\\
\hline
\end{tabular}}
\label{tab3}
\end{table}

\begin{table}[H]
\small
\centering\caption{The ResNet based network architecture and hyperparameters of LDAGAN (CelebA).}
\centering
{
\begin{tabular}{r l c c c c}
\hline
Unit & Operation & Input kernels & Output kernels & Shared\\ 
\hline
\hline
Generator& $\mathbf{z}'$ $\sim$ Normal(0, 1) & & & \\
& Linear & 257 & 1024x4x4 & No\\
& GenResBlock & 1024 & 1024 & Yes\\
& GenResBlock & 1024 & 512 & Yes\\
& GenResBlock & 512 & 256 & Yes\\
& GenResBlock & 256 & 128 & Yes\\
& GenResBlock & 128 & 64 & Yes\\
& OptimizeResBlock & 64 & 64 & Yes\\
& Conv(Tanh) & 64 & 3 & Yes\\
\hline
Discriminator & Conv & 3 & 64 &\\
& DisResBlock & 64 & 64 &\\
& DisResBlock & 64 & 128 &\\
& DisResBlock & 128 & 256 &\\
& DisResBlock & 256 & 512 &\\
& DisResBlock & 512 & 1024 &\\
& OptimizeResBlock & 1024 & 1024 &\\
& 1x1 Conv & 1024 & 1024 &\\
& Sampling & & &\\
& Linear & 1024x4x4 & 1 &\\
\hline
Number of $G$& 5& \\
Batch size of real data  &32&  \\
Batch size for each generator  &6 &\\
Learning rate  &0.0001 &\\
Optimizer  &RMSprop($0.99$)&\\
\hline
\end{tabular}}
\label{tab7}
\end{table}

\begin{table}[H]
\small
\centering\caption{ResBlock Architecture(CIFAR10, CIFAR100 and ImageNet).}
\centering
{
\begin{tabular}{r l c c r c c c c}
\hline
Unit & Operation & Kernel & Padding & Activation & BN & SN  & Resize\\ 
\hline
\hline
GenResBlock & Conv & 3x3 & 1 & Relu & Yes & No &  Upsample \\
& Conv & 3x3 & 1 & Relu & Yes & No  & None \\
\hline
OptimizeResBlock & Conv & 3x3 & 1 & Relu & No & Yes & None \\
& Conv & 3x3 & 1 & None & No & Yes &  Pooling \\
\hline
DisResBlock & Conv & 3x3 & 1 & Relu & No & Yes &  None \\
& Conv & 3x3 & 1 & Relu & No & Yes &  Pooling \\
\hline
\end{tabular}}
\label{tab4}
\end{table}

\begin{table*}
\small
\centering\caption{ResBlock architecture (CelebA).}
\centering
{
\begin{tabular}{r l c c r c}
\hline
Unit & Operation & Kernel & Padding & Activation & Resize\\ 
\hline
\hline
GenResBlock & Conv & 3x3 & 1 & LeakyRelu & None \\
& Conv & 3x3 & 1 & LeakyRelu & Upsample \\
\hline
OptimizeResBlock & Conv & 3x3 & 1 & LeakyRelu & None \\
& Conv & 3x3 & 1 & LeakyRelu & None \\
\hline
DisResBlock & Conv & 3x3 & 1 & LeakyRelu & AvgPooling \\
& Conv & 3x3 & 1 & LeakyRelu & None \\
\hline
\end{tabular}}
\label{tab:ResBlock2}
\end{table*}

\section{Parameter Sharing}
\label{App:ParShare}
We evaluated the effect of parameter sharing on the generative performance.
Inception scores (IS) and Fr\'{e}chet inception distance (FID) are two measures.
Tab.~\ref{tab:Sharing} shows the evaluation results on the CIFAR-10, CIFAR-100 and ImageNet datasets.
The results show a consistent tendency that the less the parameter sharing layers the worse the performance is.
Parameter sharing helps to keep the balance of generator's updating.
This makes it possible for discriminator to score the performance of different generators simultaneously.
This is a partial explanation of the performance dropping caused by adopting less parameter sharing layers.

\begin{table}[H]
\caption{The performances of LDAGAN (standard CNN based) using different parameter sharing schemes}
\centering
\small\addtolength{\tabcolsep}{-0pt}
\subtable[Inception scores on different datasets]{
{\begin{tabular}{lccc}
\hline
untied layer & CIFAR-10 & CIFAR-100 & ImageNet\tabularnewline
\hline
\hline
1         &  $7.46$        &        $7.57$              &    $8.34$     \tabularnewline

1-2      &  $6.13$        &         $6.910$             &    -     \tabularnewline
 \hline
\end{tabular}}
\label{tab:firstSharing}
}
\qquad
\subtable[Fr\'{e}chet inception distance on different datasets]{
{\begin{tabular}{lccc}
\hline
untied layer & CIFAR-10 & CIFAR-100 & ImageNet\tabularnewline
\hline
\hline
1         &  $24.3$        &        $28.7$             &    $28.8$     \tabularnewline

1-2      &  $44.3$        &         $47.6$             &    -     \tabularnewline
 \hline
\end{tabular}}
\label{tab:secondSharing}
}
\label{tab:Sharing}
\end{table}

\section{Different Generator Number}
\label{App:diffGen}
We tested the performances of LDAGAN with different generators on the CIFAR-10, CIFAR-100, and ImageNet datasets.
Since CIFAR-100 and ImageNet have much more image classes, more generators should be employed.
10, 20, and 30 generators were used on CIFAR-100, and 10, 20, 27 generators were used on ImageNet.
The quantitative results can be found in Tab.~\ref{tab:NumGenerator}, where the best performances are achieved when 10 and 20 generators are employed on CIFAR-100 and ImageNet respectively.
\begin{table}[H]
\caption{The performances of LDAGAN with various number of generators}
\centering
\small\addtolength{\tabcolsep}{-0pt}
\subtable[Inception scores on different datasets]{
{\begin{tabular}{lccc}
\hline
number & CIFAR-10 & CIFAR-100 & ImageNet\tabularnewline
\hline
\hline
10         &  $7.46$        &        $7.50$              &    $8.20$     \tabularnewline

20      &  -       &         $7.57$             &    $8.21$     \tabularnewline

30/27        &    -      &         $6.91$             &    $8.34$     \tabularnewline
 \hline
\end{tabular}}
       \label{tab:firstNumG}
}
\qquad
\subtable[Fr\'{e}chet Inception Distance on different datasets]{
{\begin{tabular}{lccc}
\hline
number & CIFAR-10 & CIFAR-100 & ImageNet\tabularnewline
\hline
\hline
10         &  $24.3$        &        $28.8$              &    $36.1$     \tabularnewline

20      &  -        &         $29.8$             &    $28.9$     \tabularnewline

30/27        &    -      &         $35.26$             &    $31.5$     \tabularnewline
 \hline
\end{tabular}}
\label{tab:secondNumG}
}
\label{tab:NumGenerator}
\end{table}

\section{Parameter Updating}
\label{App:ParUp}

One assumption in LDAGAN is that discriminator can output a likelihood indicating how realistic the synthesized image is.
Such a likelihood is used to calculate variational parameters $\bm{\gamma}$ and $\bm{\omega}$.
However, the discriminator is sometimes inaccurate and unstable during training.
This seems to only appears obviously early in learning.
We thus fixed the variational parameters $\bm{\gamma}$ and $\bm{\omega}$ at the beginning epochs, and kept them updating after a certain epoch.

On CIFAR-10, we tested the performance of LDAGAN when updated $\bm{\gamma}$ and $\bm{\omega}$ after 0, 50 and 100 epochs.
The corresponding inception scores are 7.36, 7.38, and 7.46, and FIDs are 26.4, 25.7, and 24.3.
These results show LDAGAN has an improved performance if we update $\bm{\gamma}$ and $\bm{\omega}$ after a certain epoch.
On ImageNet, we analyzed the FID and IS of LDAGAN when updated $\bm{\gamma}$ and $\bm{\omega}$ after 0 and 4 epochs.
The inception scores are 8.37 and 8.21, and the FIDs are 29.0 and 28.9.
There does not exist significant difference between these two strategies.

\section{Sampling Details}
\label{App:Sampling}
The learning of discriminator refers to sampling.
As we described in Sec.~4, the discriminative loss has the from of:
\begin{equation}
\label{eq:Learning-D}
\max_{\bm{\phi}}
\mathbb{E}_{\mathbf{x} \sim p_{data}\left(\mathbf{x}\right)}
\mathbb{E}\left[ \log p\left( y=1 | \mathbf{x},\bm{\phi} \right) \right]
+\mathbb{E}_{\mathbf{z}' \sim p\left(\mathbf{z}'\right)}
\mathbb{E}\left[ \log\left( 1-p\left( y=1 | \mathbf{z}',\bm{\theta},\bm{\alpha},\bm{\phi} \right) \right) \right],
\end{equation}
where $p\left( y=1 | \mathbf{z}',\bm{\theta},\bm{\alpha},\bm{\phi} \right)$ is a marginal probability, which is obtained by integrating joint distribution over $\bm{\pi}$ and summing over $\mathbf{z}$:
\begin{equation}
\int p\left(\bm{\pi} | \bm{\alpha}\right)
\left(\sum_{\mathbf{z}}
p\left(\mathbf{z}|\bm{\pi}\right)
p\left(y=1|\mathbf{z}, \mathbf{z}', \bm{\theta}, \bm{\phi}\right)
\right)d\bm{\pi},
\label{eq:Learning-AD2}
\end{equation}
$p\left(y=1|\mathbf{z}, \mathbf{z}', \bm{\theta}, \bm{\phi}\right)$ here denotes, given the underlying mode (\ie~$z_k=1$), the probability of the synthetic sample being real.
We utilize $ D\left(G_k\left(\mathbf{z}'\right)\right)$ to score this probability.
To solve $\bm{\phi}$, we should adopt ancestral sampling since the integration over $\bm{\pi}$ is analytically intractable.

Ancestral sampling, in fact, is somewhat time consuming.
On the CIFAR-10 dataset, the running time for 1 epoch with ancestral sampling is 5.10 minutes on a GTX1080Ti GPU.
To ensure the learning efficiency, we change ancestral sampling to randomly sampling fixed number of real and fake samples.
The details can be found in Tab.~\ref{tab:Sharing} and \ref{tab:NumGenerator}.
By virtue of this simplification, the running time for 1 epoch reduces to be 2.30 minutes.

\section{Generated Images}
\label{App:GenImg}

\begin{figure}[H]
  \centering
    \includegraphics[width=0.5\textwidth]{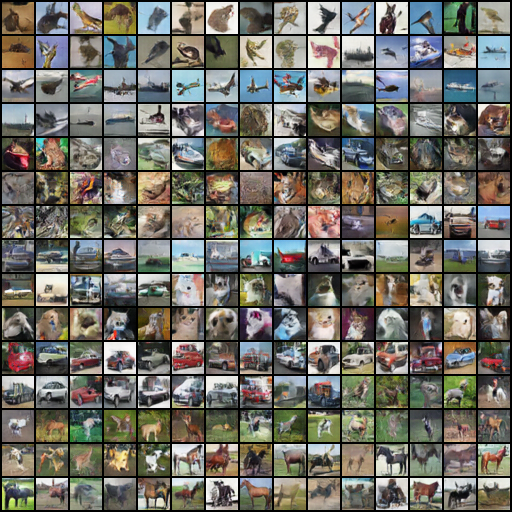}
     \caption{Images generated by LDAGAN trained on the CIFAR-10 dataset.}
     \label{fig:ExpImg1}
\end{figure}

\begin{figure}[H]
  \centering
    \includegraphics[width=0.5\textwidth]{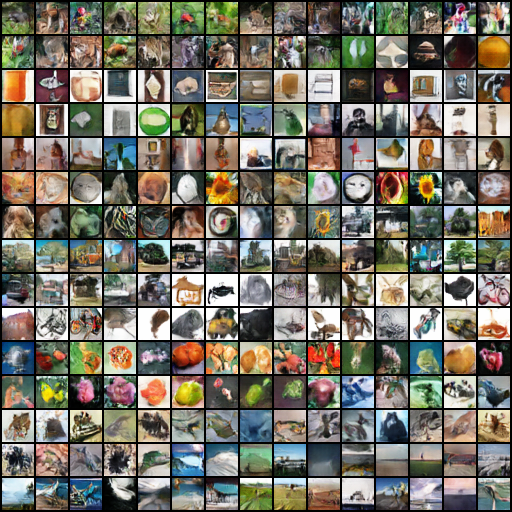}
     \caption{Images generated by LDAGAN trained on the CIFAR-100 dataset.}
     \label{fig:ExpImg2}
\end{figure}

\begin{figure}[H]
  \centering
    \includegraphics[width=0.65\textwidth]{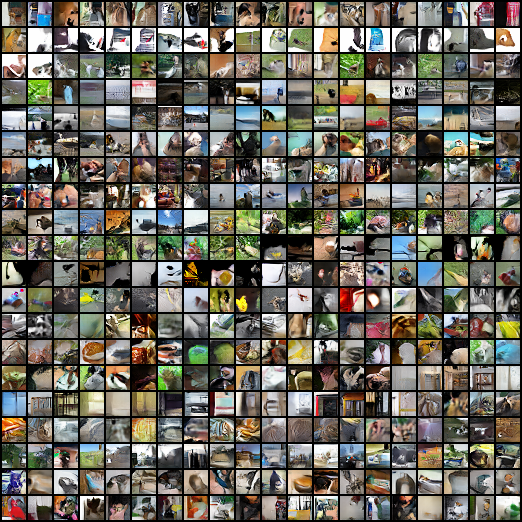}
     \caption{Images generated by LDAGAN trained on the rescaled $32\times32$ ImageNet dataset.}
     \label{fig:ExampImg3}
\end{figure}

Some example images generated by LDAGAN (with CNN architecture) trained on the CIFAR-10, CIFAR-100 and ImageNet datasets are shown in Fig.~\ref{fig:ExpImg1}, Fig.~\ref{fig:ExpImg2}, and Fig.~\ref{fig:ExampImg3}, respectively.

\section{High Resolution Image Generation}

\begin{figure}[t]
  \centering
    \includegraphics[width=0.8\textwidth]{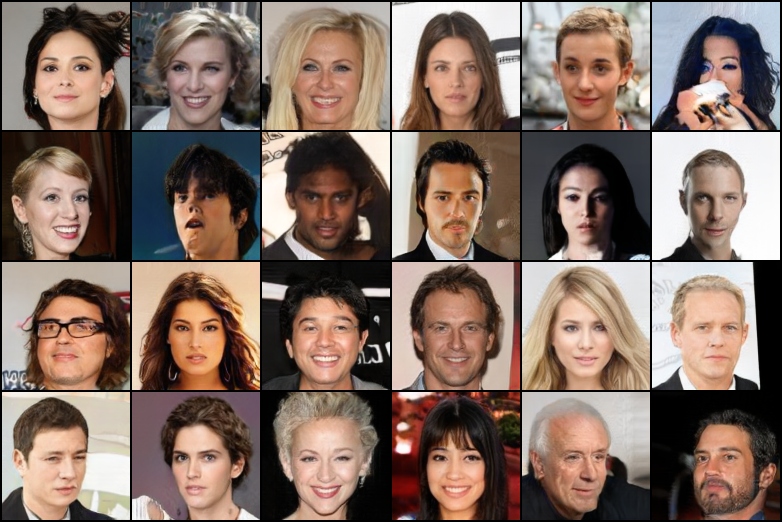}
     \caption{Random LDAGAN samples on CelebA $128\times128$.}
     \label{fig:fake}
\end{figure}

On CelebA dataset, we generated images of size $128\times 128$.
Fig.~\ref{fig:fake} shows the random LDAGAN samples on CelebA.
The network architecture for generating the high resolution images is described in Tab.~\ref{tab7}

\end{document}